\documentclass[sigconf]{acmart}
\usepackage{multirow}
\usepackage{subfig}

\usepackage{algorithm}
\usepackage{algorithmic}
\usepackage{amsmath}
\usepackage{enumitem}
\usepackage{balance}
\AtBeginDocument{%
  }

\copyrightyear{2025}
\acmYear{2025}
\setcopyright{acmlicensed}
\acmConference[CIKM '25] {Proceedings of the 34th ACM International Conference on Information and Knowledge Management}{ November 10--14, 2025}{Seoul, Republic of Korea.}
\acmBooktitle{Proceedings of the 34th ACM International Conference on Information and Knowledge Management (CIKM '25), November 10--14, 2025, Seoul, Republic of Korea}
\acmISBN{979-8-4007-2040-6/2025/11}
\acmDOI{10.1145/3746252.3761005}

\begin{document}

\title{Exploring the Tradeoff Between Diversity and Discrimination \\for Continuous Category Discovery}

\author{Ruobing Jiang}
\orcid{0000-0003-1209-078X}
\affiliation{%
  \institution{Ocean University of China}
  \city{Qingdao}
  \country{China}}
\email{jrb@ouc.edu.cn}

\author{Yang Liu}
\orcid{0009-0002-1919-7678}
\affiliation{%
  \institution{Ocean University of China}
  \city{Qingdao}
  \country{China}}
\email{ly6611@stu.ouc.edu.cn}

\author{Haobing Liu}
\orcid{0000-0002-2546-3306}
\authornote{Haobing Liu is the corresponding author.}
\affiliation{%
  \institution{Ocean University of China}
  \city{Qingdao}
  \country{China}}
\email{haobingliu@ouc.edu.cn}

\author{Yanwei Yu}
\orcid{0000-0002-5924-1410}
\affiliation{%
  \institution{Ocean University of China}
  \city{Qingdao}
  \country{China}}
\email{yuyanwei@ouc.edu.cn}

\author{Chunyang Wang}
\orcid{0000-0002-1752-5423}
\affiliation{%
  \institution{East China Normal University}
  \city{Shanghai}
  \country{China}}
\email{cywang@dase.ecnu.edu.cn}

\renewcommand{\shortauthors}{Ruobing Jiang, Yang Liu, Haobing Liu, Yanwei Yu, \& Chunyang Wang}

\begin{abstract}
Continuous category discovery (CCD) aims to automatically discover novel categories in continuously arriving unlabeled data.
This is a challenging problem considering that there is no number of categories and labels in the newly arrived data,  while also needing to mitigate catastrophic forgetting.
Most CCD methods cannot handle the contradiction between novel class discovery and classification well. They are also prone to accumulate errors in the process of gradually discovering novel classes. Moreover, most of them use knowledge distillation and data replay to prevent forgetting, occupying more storage space.
To address these limitations, we propose Independence-based Diversity and Orthogonality-based Discrimination (IDOD).
IDOD mainly includes independent enrichment of diversity module, joint discovery of novelty module, and continuous increment by orthogonality module.
In independent enrichment, the backbone is trained separately using contrastive loss to avoid it focusing only on features for classification.
Joint discovery transforms multi-stage novel class discovery into single-stage, reducing error accumulation impact.
Continuous increment by orthogonality module generates mutually orthogonal prototypes for classification and prevents forgetting with lower space overhead via representative representation replay.
Experimental results show that on challenging fine-grained datasets, our method outperforms the state-of-the-art methods.
\end{abstract}



\maketitle

\begin{figure}[t]
    \centering
    \includegraphics[scale=1.0]{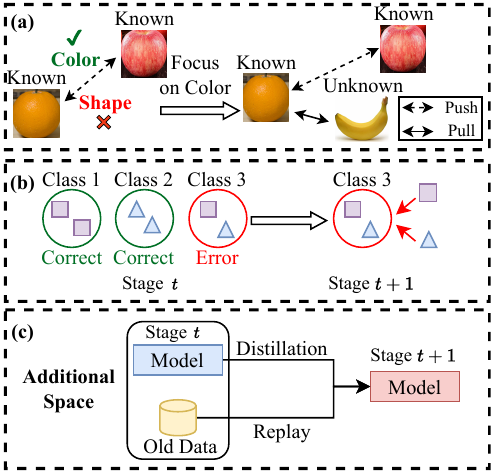}
    \caption{Challenges in CCD. In Figure~\ref{fig:Intro} (a), for orange and apple, model tends to select color feature rather than shape feature. Focusing only on color leads to difficulty in distinguishing between orange and banana. In Figure~\ref{fig:Intro} (b), the error in discovering novel classes in previous stage affect the assignment of pseudo-labels in subsequent stages. In Figure~\ref{fig:Intro} (c), model retaining known class knowledge and known class data are stored to prevent catastrophic forgetting.}
    \label{fig:Intro} 
\end{figure}

\section{Introduction}

Deep learning methods have outstanding performance in image classification tasks~\cite{deepintro,cvcikm}.
However, most methods are still under the closed-set assumption, which means that the classes in the test set are consistent with those in the training set.
The assumption limits the practicability of these methods~\cite{introopen}. Considering that in the real world, there is a need to handle continuously arriving unlabeled data, which contains novel classes.
It is extremely meaningful to explore methods that can adapt to the real world~\cite{openworld,open2,open3,opencikm}.

A large number of methods such as novel category discovery (NCD) methods~\cite{IJCAI-NCD,introNCD}, and continuous category discovery (CCD) methods~\cite{GMCCD,CGCD2025} have been proposed to deal with scenarios where novel classes appear.
However, NCD methods do not consider the situation of data distribution shift caused by the continuous arrival of data.
CCD methods are more practical than NCD methods, considering that they can discover novel classes in continuously arriving unlabeled data.

Although existing CCD methods have achieved certain success, there are still three challenges that need to be addressed, as shown in Figure~\ref{fig:Intro}.
During classification, CCD methods are inclined to select the features that are most crucial for classification and discard some other features.
When discovering novel classes, features that are effective for classification may lead to an inability to distinguish novel classes from old ones~\cite{closer}.
As Figure~\ref{fig:Intro}(a) shows, these two different requirements act on model training simultaneously, making it difficult for the features to achieve a better balance between diversity and discrimination.
\textbf{So the first challenge is how to handle the different requirements for features between discovering novel classes and classification}.
Existing CCD methods determine the number of novel classes from unlabeled data at each stage and assign pseudo-labels to unlabeled data. As Figure~\ref{fig:Intro}(b) shows, these pseudo-labels are not entirely accurate. After processing multiple stages, errors will accumulate.
\textbf{So the second challenge is how to reduce the error accumulation of discovering novel classes in unlabeled data.}
Most CCD methods use knowledge distillation~\cite{PACCD} or/and data replay~\cite{GMCCD} to address catastrophic forgetting. These methods require storing the models and data from the previous stages, occupying more storage space~\cite{space}.
\textbf{So the third challenge is how to address catastrophic forgetting with lower space overhead.}

To address the above challenges, \textbf{I}ndependence-based \textbf{D}iversity and \textbf{O}rthogonality-based \textbf{D}iscrimination (\textbf{IDOD} in short) with a backbone that generates diverse features and a projector that extracts discriminative features was proposed.

\textbf{To solve the first challenge}, independent enrichment of diversity module first trains the backbone and prototypes using contrastive loss. Backbone is used to generate diverse features. Then, the parameters of the backbone are fixed to avoid being affected by subsequent training. In this way, the backbone focuses on more diverse features rather than being limited to those used for classification. Prototypes are used to select novel samples.

\textbf{To solve the second challenge}, joint discovery of novelty module uses the frozen backbone to process multi-stage novel class samples. The representations of novel class samples that appeared in previous stages are stored in a dynamic pool. The representations of novel class samples in the current stage are combined with the representations in the dynamic pool and used for discovering novel classes.
This module transforms multiple stages into a single stage, minimizing the impact of error accumulation.
On the one hand, joint discovery of novelty module avoids the adverse effects of errors in the previous stage on the subsequent stage. On the other hand, this approach can combine samples from multiple stages to enrich the sample quantity.

\textbf{To solve the third challenge}, mutually orthogonal prototypes are generated for each class in the feature space.
Considering that the backbone is inclined to extract diverse features, a projector is introduced to extract discriminative features.
The projector is trained with the assistance of these prototypes. At the same time, forgetting is prevented by replaying the representations of representative known class samples.
In this way, the best positions in the feature space for all classes are reserved. In addition, the representations rather than the raw data are utilized for replay. These methods can prevent forgetting with a smaller space overhead.

The main contributions of the method proposed in this paper can be summarized as follows:

\newcommand{\customitem}[1]{%
  \noindent\makebox[1.5em][l]{\scalebox{2.0}{\textbullet}}%
  \begin{minipage}[t]{\dimexpr\linewidth - 1.5em\relax}#1\end{minipage}\par
}
\customitem{Independent enrichment of diversity module that can better handle the tradeoff between diversity and discrimination in continuous category discovery is proposed.}
\customitem{Joint discovery of novelty module that can minimize the impact of error accumulation caused by successive discovery of novel classes is proposed.}
\customitem{Continuous increment by orthogonality module that can prevent forgetting with lower space overhead is proposed.}
\customitem{The proposed method performs better than state-of-the-art methods on multiple fine-grained datasets.}

\section{Related Work}

The research directions related to this paper include class-incremental learning (CIL), novel category discovery (NCD), and continuous category discovery (CCD).
However, CIL and NCD cannot fit real-world scenarios very well, and
existing CCD methods have limitations in discovering novel classes and preventing forgetting.

\textbf{Class-incremental learning} aims to learn novel classes from continuously arriving data while not forgetting known classes~\cite{CILaddition1}.
However, in CIL, the assumption that continuously arriving data is labeled limits its application in real-world scenarios~\cite{CIL,introCIL}.
Most CIL methods are dedicated to addressing catastrophic forgetting. Knowledge distillation~\cite{CILKNOWLEDGE1,CILKNOWLEDGE2,CILKNOWLEDGE3,CILKNOWLEDGE4}, data replay~\cite{CILDATA1,CILDATA2,CILDATA3,CILDATA4}, and weight consolidation~\cite{CILWEI1,CILWEI2} are commonly used.
To enhance the practicality of CIL methods, a large number of few-shot class-incremental learning methods have been proposed~\cite{orco,FSCIL}.

\textbf{Novel category discovery} aims to automatically discover novel classes from unlabeled data~\cite{NCDaddition1,NCDaddition2}. It is more in line with real scenarios than CIL.
However, NCD does not consider the situation where data arrives continuously and most NCD methods require prior knowledge of the number of novel classes. This makes NCD still have limitations that make it not applicable to real scenarios~\cite{CILNCD}.
NCD can be divided into two categories according to whether the unlabeled data they process contains known classes.
One type is that there are only novel classes in the unlabeled data~\cite{NCD11,NCD12,NCD13,NCD14}. The other type is that there are both novel classes and known classes in the unlabeled data~\cite{NCD21,NCD22,NCD23}.

\begin{figure*}[t]
    \centering
    \includegraphics[scale=0.37]{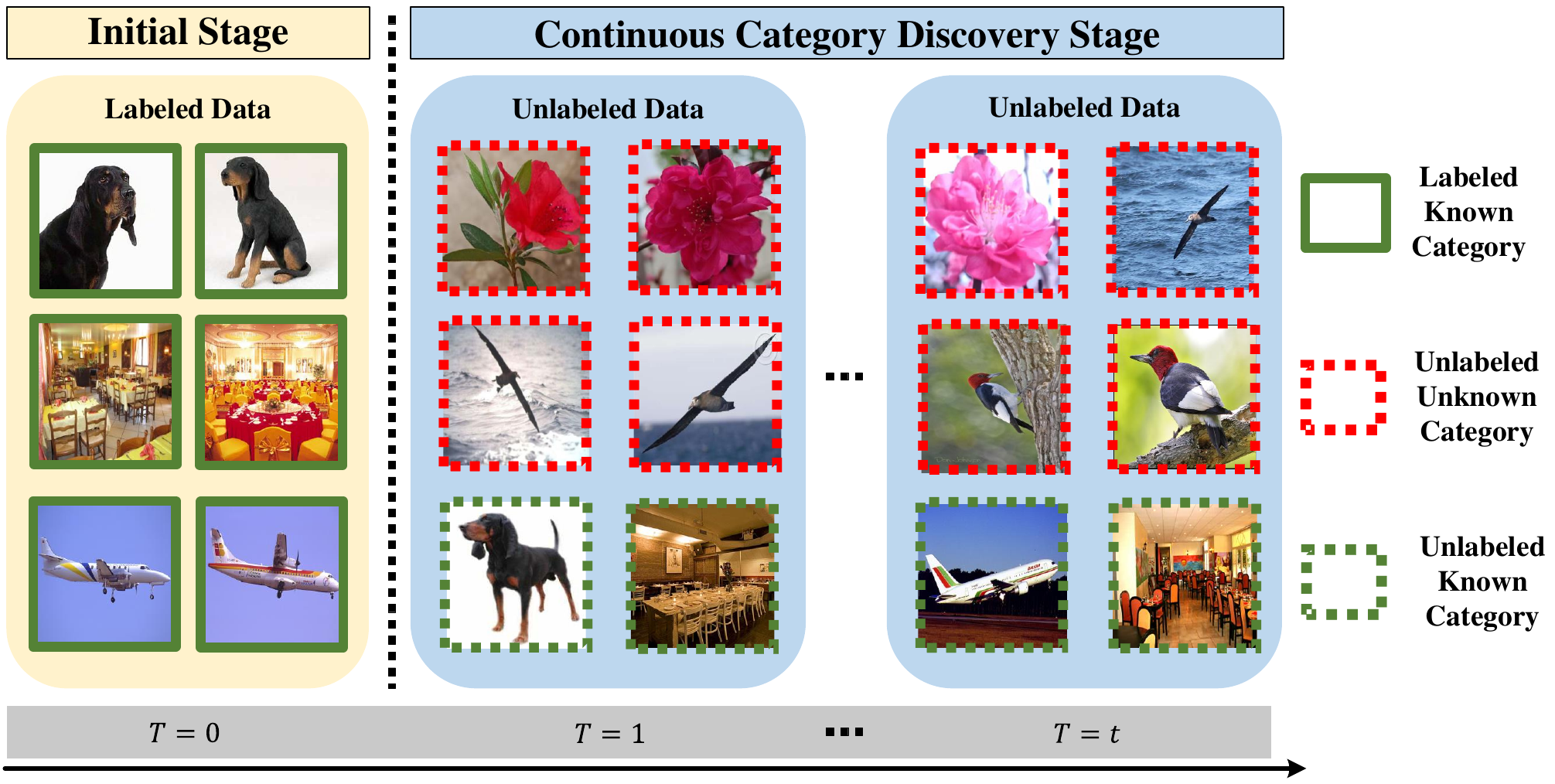}
    \caption{Continuous category discovery task.}
    \label{fig:CCD} 
\end{figure*}

\textbf{Continuous category discovery} aims to automatically discover novel classes in continuously arriving unlabeled data without the conditions of labels and the number of categories, and maintain the recognition performance for known classes.
CCD is more in line with real scenarios than both CIL and NCD.
However, most CCD methods have deficiencies in finding novel classes and preventing forgetting.
To find novel classes, the model needs to be able to extract rich and diverse features. To accurately classify, the model needs to be able to extract key discriminative features.
In existing CCD methods, these two contradictory requirements will act on the training of the backbone simultaneously.
Considering the absence of labels and the number of categories, for each batch of unlabeled data, there will be errors in the results of finding novel classes. Existing CCD methods will accumulate errors in the case of continuous arrival of data.
Furthermore, most CCD methods employ knowledge distillation and data replay to prevent forgetting. Additional model and data assist model in retaining the knowledge of known classes at a greater space cost.
GM~\cite{GMCCD} designs a growth stage and a merging stage. In growth stage, novel classes are found. In merging stage, the knowledge of known and novel classes is distilled into the same model.
Furthermore, there are CCD methods based on proxy-anchor method~\cite{PACCD}, meta-learning method~\cite{METACCD}, and prompt learning method~\cite{PROMPTCCD}.

\begin{table}[t]
  \centering
  \caption{Notations and descriptions}
  \renewcommand{\arraystretch}{1.5}
  \renewcommand{\heavyrulewidth}{1pt}
  \label{tab:notations}
\begin{tabular}{ll}
\toprule
\textbf{Notations} & \textbf{Descriptions}               \\ \midrule
$D^l$                & Labeled data \\
$D^u_i$                & Unlabeled data at stage $i$                 \\
$p$                & Prototype \\
$g$                & Orthogonal prototype                 \\

$\epsilon$                & Threshold used for non-parametric split                 \\
$S^\text{o}$                & Static pool   \\

$S^\text{n}_i$       & Dynamic pool at stage $i$       \\

$F_\theta$                & Backbone   \\

$F_p$       & Projector      \\
$f$                & Output of the backbone   \\

$z$       & Output of the projector       \\
\bottomrule
\end{tabular}
\end{table}

\section{Proposed Method}
\subsection{Problem Definition and Method Overview}
\subsubsection{Problem Definition}
There are many problem settings that take into account the emergence of novel classes in data.
However, compared with CCD, they all have different limitations, which makes it impossible for them to better adapt to the real environment.
The important notations and their corresponding descriptions are presented in Table~\ref{tab:notations}.

CCD consists of two phases, namely the initial phase and the continuous category discovery phase, as shown in Figure~\ref{fig:CCD}.
In the initial phase, labeled training data $D^l$ is used to train the model.
The classes that appear in $D^l$ are known classes.
The model is expected to have the ability to classify known classes.
In the continuous category discovery phase, the model needs to process a series of unlabeled data $D^u_t$, where $D^u_t$ represents the data at stage $t$.
$D^u_t$ contains both known classes and unknown classes.
The model needs to discover novel classes from $D^u_t$ when the number of novel classes is unknown.
Ultimately, the model should possess the ability to classify both known and unknown classes.

\begin{figure*}[t]
    \centering
    \includegraphics[scale=0.8]{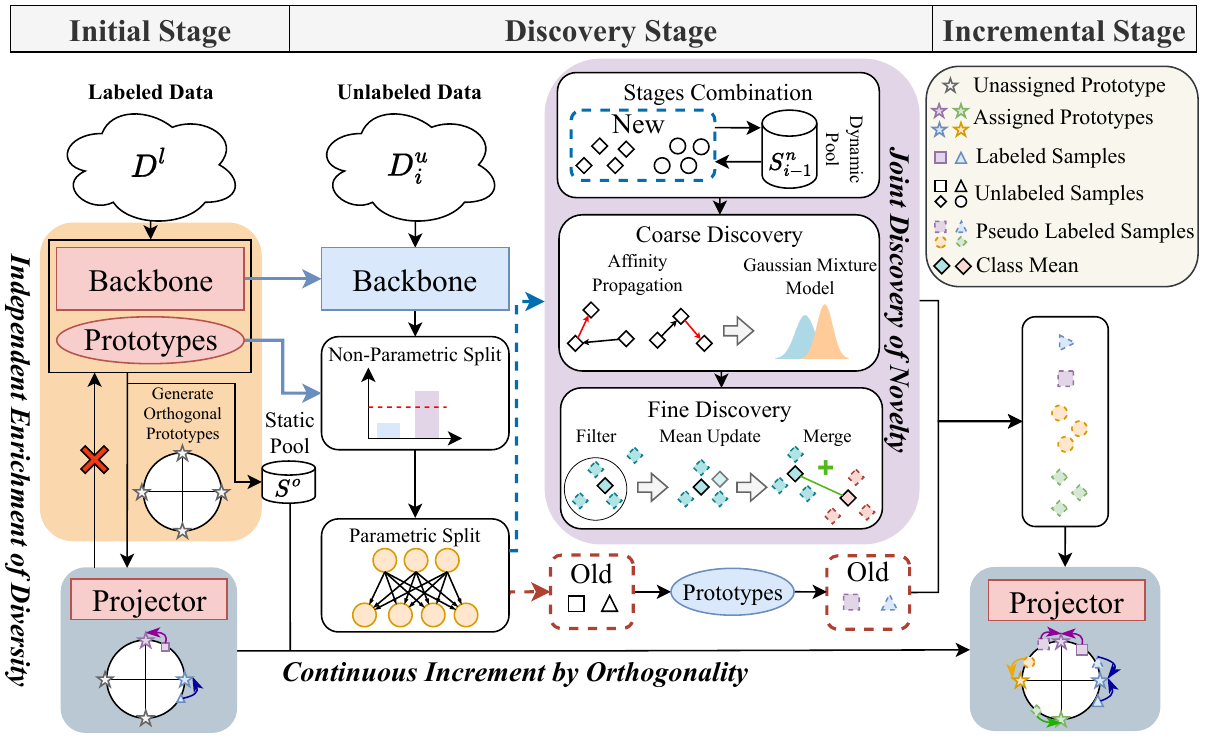}
    \caption{Overview of IDOD framework. According to the problem setting of CCD, IDOD includes the initial stage, discovery stage and incremental stage to process data. In initial stage, independent enrichment of diversity module separately trains backbone and projector. This enables backbone to focus on diverse features and projector to focus on discriminative features. In discovery stage, joint discovery of novelty module transforms multiple stages into a single stage, reducing the error of novel class discovery, considering that it can decrease error accumulation and enrich the sample size in the subsequent stage. In incremental stage, the orthogonal prototypes in continuous increment by orthogonality module are used to assist in the learning of novel classes.}
    \label{fig:IDOD} 
\end{figure*}

\subsubsection{Method Overview}
Figure~\ref{fig:IDOD} shows an overview of IDOD. IDOD mainly includes independent enrichment of diversity module, joint discovery of novelty module, and continuous increment by orthogonality module.
The labeled data is used to train the backbone and the projector. Moreover, the trainings of the backbone and the projector are independent of each other.
Subsequently, the backbone is frozen and used to extract the representations of unlabeled data.
Unlabeled data is divided into novel classes and known classes through the split operation. They obtain corresponding pseudo-labels through joint discovery and prototypes, respectively.
Orthogonal prototypes assign a prototype to each class. The projector is trained through cross-entropy.

Backbone, projector, and orthogonal prototypes are used to predict the labels of test samples.
The representations of samples are obtained by successively using the backbone and the projector.
Then, the similarity between the sample representations and the orthogonal prototypes is calculated.
The class label corresponding to the orthogonal prototype with the highest similarity is the predicted label of the sample.

\subsection{Independent Enrichment of Diversity}
Independent enrichment of diversity module trains the backbone independently to maintain the ability of the backbone to generate diverse features.
In this way, the backbone is enabled to focus on more diverse and rich features, while the projector is allowed to concentrate on key discriminative features.

For the training of the backbone, firstly, a prototype is generated for each known class. These prototypes are randomly generated. $P$ denotes the set of these prototypes.
According to the objective of maximizing the similarity between the sample and the corresponding prototype while minimizing the similarity with other prototypes. The backbone and the prototypes are trained together. For sample $x_i$, the loss function is defined as the following equation:

\begin{equation}
\mathcal{L}^i_\text{CL} = -\alpha(s(f_i,p^+)-\sigma)+\frac{1}{|P^-|}\sum_{p\in P^-}\alpha(s(f_i,p)+\sigma),
\label{CL} 
\end{equation}
where $\alpha$ is a scaling factor, $s$ represents cosine similarity. $f_i$ is the sample representation obtained from the backbone, $p^+$ is the prototype corresponding to the current sample, $P^-$ is the set of all other prototypes, and $\sigma$ is used to control the boundary value of similarity. $\alpha$ and $\sigma$ are hyperparameters

After the completion of the aforementioned training, to avoid the impact of subsequent training on the backbone, the backbone is frozen.
On the one hand, the contrastive loss helps the backbone pay attention to more general representations.
On the other hand, the frozen backbone has the ability to prevent forgetting.
The representation obtained by passing the data through the backbone is used as input for the training of the projector.
In addition, to achieve replay, for each known class, the $m$ representative representations closest to the class mean are stored in the static pool according to the following equation:

\begin{equation}
S_c^\text{o} = \{f_i|f_i\in \mathcal{T}^m(\text{distance}(f_i,f_c))\},
\label{topm} 
\end{equation}
where $S_c^\text{o}$ represents the set of representative representations of $c$, $f_c$ is the mean value of all sample representations of category $c$ and $\text{distance}$ is euclidean distance. Static pool can be used to prevent catastrophic forgetting.

For the training of the projector, in order to make it focus on key discriminative features, orthogonal prototypes will be used to assist in its training. The specific steps will be described in detail in Sec.~\ref{4.3}.

\subsection{Joint Discovery of Novelty}
Joint discovery of novelty module includes sample union, coarse discovery and fine discovery.
It transforms the novel class discovery in multiple stages into a single stage.
On the one hand, this can avoid the inaccurate novel class discovery results of the current stage from having an impact on subsequent novel class discovery.
On the other hand, this can make the sample size of each novel class more abundant, which helps to improve the results of novel class discovery.

Before joint discovery, unlabeled data will be divided into novel classes and known classes through non-parametric split and parametric split.
The similarity between samples and the prototypes of all known classes is calculated. Based on a threshold, samples can be classified into novel classes and known classes.
In the above split results, the part with the lowest similarity in novel class data and the part with the highest similarity in known class data are used as reliable training samples $S_\text{MLP}$ for the training of MLP as follows:

\begin{equation}
S_\text{MLP} = \{f_i|\max_{p\in P}(s(f_i,p))\leq \epsilon-\delta||\max_{p\in P}(s(f_i,p))\geq \epsilon+\delta\},
\label{filter} 
\end{equation}
where $P$ is the set of prototypes of known classes, $\epsilon$ is a threshold and $\delta$ is a fixed value used to select more reliable samples.

In stages combination, representations of novel class data from the previous stage are stored through a dynamic pool to achieve the combination of the representations from the previous stage and the current stage.
Thanks to freezing the backbone, the representation obtained by passing the data through the backbone remains unchanged.
The novel class data that has appeared in the past can be stored in the form of representations and used together with the representations of the current novel class data for novel class discovery.
The update of the dynamic pool is as follows:

\begin{equation}
S^\text{n}_i = S^\text{n}_{i-1} \cup F^\text{n}_i,
\label{pool} 
\end{equation}
where $S^\text{n}_i$ is the dynamic pool at the current stage, $S^\text{n}_{i-1}$ is the dynamic pool at the previous stage, and $F^\text{n}_i$ is the set of novel class sample representations at the current stage.

For the novel class representations obtained from above steps, pseudo-labels will be assigned to them by coarse discovery and fine discovery.
In coarse discovery, Affinity Propagation~\cite{AP} is first used to estimate the number of novel classes, and then a Gaussian mixture model is used for clustering according to the number of classes.
For each novel class obtained in coarse discovery.
Samples with higher confidence are selected as the novel class discovery results.
In fine discovery, for each novel class, $k$ samples closest to the class mean are selected and the class mean is updated according to the following equation:
\begin{equation}
\overline{f_c} = \frac{1}{k}\sum_{i=1}^kf_{c,i},
\label{junzhi} 
\end{equation}
where $f_{c,i}$ is the $i$-th representation of $c$.

In addition, to make the number of novel classes more reasonable, a dynamic threshold is used to merge novel classes.
According to the distance between class means, classes with a distance lower than the threshold are merged into one class.
This threshold is obtained through known class data.
Process known class data using rough discovery and fine discovery.
Try different thresholds and select the threshold for which the number of merged categories is closest to the true number of known classes.
The formula for obtaining the threshold is shown in the following equation:

\begin{equation}
\lambda = \arg\min_{\lambda}\left| \left| C_\lambda \right|- \left| C_0 \right| \right|,
\label{yuzhi} 
\end{equation}
where $|C_0|$ is the number of known classes, and $|C_\lambda|$ is the number of known classes estimated with a threshold of $\lambda$.

This approach enables dynamic adjustment of the threshold for different datasets. In addition, novel classes can be discovered with a granularity similar to that of the known classes.

\subsection{Continuous Increment by Orthogonality}\label{4.3}

\begin{algorithm}[t]
	\caption{Training Process of IDOD} 
	\label{alg3} 
        \renewcommand{\algorithmicrequire}{\textbf{Input:}}
        \renewcommand{\algorithmicensure}{\textbf{Output:}}
	\begin{algorithmic}[1]
		\REQUIRE Labeled Data $D^l$, unlabeled data $D^u_i$
		\ENSURE Backbone $F_\theta$, projector $F_p$, prototypes $P$, orthogonal prototypes $G$ and static pool $S^o$\\
            \STATE Pretrain $F_\theta$, initialize $F_p$, randomly initialize $P$ and $G$
		\WHILE{$\text{epoch}_1$} 
            \STATE $x \gets \text{Sample}(D^l)$
            \STATE $f \gets F_\theta(x)$
            \STATE $\mathcal{L}_\text{CL} \gets \mathcal{L}_\text{CL}(f,P)$ \hfill $\triangleright$ Eq. ~\eqref{CL}
            \STATE $f_\theta,P \gets \text{Adam}(\mathcal{L}_\text{CL})$,~$\text{epoch}_1 \gets \text{epoch}_1 -1$
		\ENDWHILE
            \STATE $S^o \gets \text{topk}(f)$ \hfill $\triangleright$ Eq. \eqref{topm}
            \WHILE{$\text{epoch}_2$}
            \STATE $z \gets F_p(f)$
            \STATE $\mathcal{L}_\text{CE} \gets \mathcal{L}_\text{CE}(z,G)$ \hfill $\triangleright$ Eq. ~\eqref{jiaocha}
            \STATE $F_p \gets \text{Adam}(\mathcal{L}_\text{CE})$,~$\text{epoch}_2 \gets \text{epoch}_2 -1$
            \ENDWHILE
            \STATE $f^\prime \gets F_\theta(x^\prime)$, $x^\prime \gets \text{Sample}(D^u_i)$
            \STATE $f_\text{new} \gets \text{MLP}(f^\prime)$ \hfill $\triangleright$ Eq. ~\eqref{filter}
            \STATE Assign pseudo-labels \hfill $\triangleright$ Eq. ~\eqref{pool}-\eqref{yuzhi}
            \WHILE{$\text{epoch}_3$} 
            \STATE $z_\text{new} \gets F_p(f_\text{new})$
            \STATE $\mathcal{L}_\text{CE} \gets \mathcal{L}_\text{CE}(z_\text{new},G)$ \hfill $\triangleright$ Eq. ~\eqref{jiaocha}
            \STATE $\mathcal{L}_\text{static} \gets \mathcal{L}_\text{static}(S^\text{o},G)$ \hfill $\triangleright$ Eq. ~\eqref{replay}
            \STATE $F_p \gets \text{Adam}(\mathcal{L}_\text{CE} + \mathcal{L}_\text{static})$ \hfill $\triangleright$ Eq. ~\eqref{incre}
            \STATE $\text{epoch}_3 \gets \text{epoch}_3 -1$
            \ENDWHILE
	\end{algorithmic} 
\end{algorithm}

Continuous increment by orthogonality module assigns the best positions in the feature space to all known and unknown classes to assist the projector in focusing on discriminative features.
At the same time, combined with representative representation replay, it prevents forgetting with lower space overhead.

First, a large number of prototypes are randomly generated, and the number of generated prototypes is determined by the dimension of the representation.
Specify the number of prototypes according to the dimension. Considering that at most vectors of this number can be generated and ensuring that they are orthogonal to each other.
These prototypes are made orthogonal to each other by the following equation:

\begin{equation}
\mathcal{L}_\text{G} = \frac{1}{|d|}\sum_i^{|d|}\log\sum_{j=1}^{|d|}\exp(\frac{g_i\cdot g_j}{\tau}),
\label{zhengjiao} 
\end{equation}
where $|d|$ is the dimension of $g$, $g_i$ is the $i$-th orthogonal prototype, and $\tau$ is a temperature parameter.

Being orthogonal to each other in the feature space means that these prototypes are as far away from each other as possible.
With the assistance of orthogonal prototypes, it helps the projector to focus on the discriminative features that are crucial for classification.

Then calculate the class mean of each class and assign this class label to the unassigned prototype with the largest cosine similarity, as shown in the following equation:

\begin{equation}
g_i = \arg\max_{g_i\in G^u}(s(\overline{f_c},g_i)),
\label{assign} 
\end{equation}
where $G^u$ is the set of unassigned orthogonal prototypes, and $\overline{f_c}$ is the mean of sample representation.
After the assignment is completed, cross-entropy is used to train the projector.
When training the projector, the loss function is shown in the following equation:

\begin{equation}
\mathcal{L}_\text{CE} = -\sum^N_iy_i\log\frac{\exp(z_i\cdot g^\top_i)}{\sum^{|C_0|}_{k=1}\exp(z_i\cdot g^\top_k)},
\label{jiaocha} 
\end{equation}
where $N$ is the number of training samples. $z_i$ is the vector obtained by passing sample $x_i$ through the backbone and projector. $g_i$ is the corresponding prototype.

On the one hand, orthogonal prototypes are helpful for distinguishing different classes.
On the other hand, orthogonal prototypes always retain the prototypes corresponding to known classes, which helps to reduce the forgetting caused by continuously learning novel classes.
In incremental stage, representative representations are also used to train the projector. The loss function is defined as the following equation:

\begin{equation}
\mathcal{L}_\text{static} = -\sum^{|C_0|\cdot k_0}_iy_i\log\frac{\exp(F_p(f_i)\cdot g^\top_i)}{\sum^{|C|}_{k=1}\exp(F_p(f_i)\cdot g^\top_k)},
\label{replay} 
\end{equation}
where $k_0$ is the number of representations saved for each known class, and $F_p$ is the projector.
The total loss function of the incremental stage is the following equation:

\begin{equation}
\mathcal{L}_\text{all} = \mathcal{L}_\text{CE} + \mathcal{L}_\text{static}.
\label{incre} 
\end{equation}

The training process of IDOD is provided in Algorithm~\ref{alg3}.
Line 2-9 and line 10-15 respectively correspond to the training of the backbone and the projector in independent enrichment of diversity module.
Line 16-25 correspond to the training of the projector in continuous increment by orthogonality module.

\begin{table}[t]
  \centering
  \caption{The number of classes in different training data.}
  \renewcommand{\arraystretch}{1.5}
  \renewcommand{\heavyrulewidth}{1pt}
	\begin{tabular}{crrrr}
		\toprule
		\textbf{Data} & \multicolumn{1}{c}{\textbf{CUB200}} & \multicolumn{1}{c}{\textbf{MIT67}} & \multicolumn{1}{c}{\textbf{DOG}} & \multicolumn{1}{c}{\textbf{AIR}} \\ \hline
		$D^l$    & 140                        & 46                        & 84                      & 70                      \\
		$D^u_1$    & 160                        & 53                        & 96                      & 80                      \\
		$D^u_2$    & 180                        & 60                        & 108                     & 90                      \\
		$D^u_3$    & 200                        & 67                        & 120                     & 100                     \\ \bottomrule
	\end{tabular}
  \label{tab:The number of classes in different training data}
\end{table}

\begin{table}[t]
  \centering
  \caption{Data splits.}
  \renewcommand{\arraystretch}{1.5}
  \renewcommand{\heavyrulewidth}{1pt}
	\begin{tabular}{crrrr}
		\toprule
		\textbf{Classes} & \multicolumn{1}{c}{$t=0$} & \multicolumn{1}{c}{$t=1$} & \multicolumn{1}{c}{$t=2$} & \multicolumn{1}{c}{$t=3$} \\ \hline
		[1, $0.7*|C|$]   & 87\%                   & 7\%                    & 3\%                    & 3\%                    \\
		($0.7*|C|$, $0.8*|C|$]   & 0\%                    & 70\%                   & 20\%                   & 10\%                   \\
		($0.8*|C|$, $0.9*|C|$]   & 0\%                    & 0\%                    & 90\%                   & 10\%                   \\
		($0.9*|C|$, $|C|$]   & 0\%                    & 0\%                    & 0\%                    & 100\%                  \\ \bottomrule
	\end{tabular}
  \label{tab:Data splits}
\end{table}

\section{Experiments}
In this section, Sec.~\ref{5.1} provides a detailed introduction to the datasets, evaluation metrics, and implementation details.
Sec.~\ref{5.2} presents a comparison with state-of-the-arts methods in terms of recognition, category number estimation, and space overhead.
Sec.~\ref{5.3} introduces ablation experiments used to verify the effectiveness of each component.
Sec.~\ref{5.4} analyzes each hyperparameter.

\begin{table*}[t]
  \centering
  \caption{Comparison results with other methods on four datasets.}
  \renewcommand{\arraystretch}{1.5}
  \renewcommand{\heavyrulewidth}{1pt}
  \setlength{\tabcolsep}{2pt}
	\begin{tabular}{crrrcrrrcrrrcrrr}
		\toprule
		\multirow{2}{*}{\textbf{Method}} & \multicolumn{3}{c}{\textbf{CUB200}}                                                  & \multirow{7}{*}{} & \multicolumn{3}{c}{\textbf{MIT67}}                                                   & \multirow{7}{*}{} & \multicolumn{3}{c}{\textbf{Dogs}}                                                    & \multirow{7}{*}{} & \multicolumn{3}{c}{\textbf{Air}}                                                     \\ \cline{2-4} \cline{6-8} \cline{10-12} \cline{14-16}
		                        & \multicolumn{1}{c}{\textbf{$\mathcal{M}_\text{o}\uparrow$}} & \multicolumn{1}{c}{\textbf{$\mathcal{M}_\text{f}\downarrow$}} & \multicolumn{1}{c}{\textbf{$\mathcal{M}_\text{d}\uparrow$}} &                   & \multicolumn{1}{c}{\textbf{$\mathcal{M}_\text{o}\uparrow$}} & \multicolumn{1}{c}{\textbf{$\mathcal{M}_\text{f}\downarrow$}} & \multicolumn{1}{c}{\textbf{$\mathcal{M}_\text{d}\uparrow$}} &                   & \multicolumn{1}{c}{\textbf{$\mathcal{M}_\text{o}\uparrow$}} & \multicolumn{1}{c}{\textbf{$\mathcal{M}_\text{f}\downarrow$}} & \multicolumn{1}{c}{\textbf{$\mathcal{M}_\text{d}\uparrow$}} &                   & \multicolumn{1}{c}{\textbf{$\mathcal{M}_\text{o}\uparrow$}} & \multicolumn{1}{c}{\textbf{$\mathcal{M}_\text{f}\downarrow$}} & \multicolumn{1}{c}{\textbf{$\mathcal{M}_\text{d}\uparrow$}} \\ \hline
		GM\cite{GMCCD}                      & 10.93                   & 41.64                 & 9.79                    &                   & 18.67                   & 40.63                 & 17.15                   &                   & 7.93                    & 48.61                 & 6.17                    &                   & 13.62                   & 51.25                 & 15.72                   \\
		MetaGCD\cite{METACCD}                 & 59.27                   & 15.31                 & 20.41                   &                   & 49.22                   & 21.60                 & 20.18                   &                   & 69.29                   & 11.40                 & 23.71                   &                   & 30.27                   & 21.63                 & 19.58                   \\
		PA-CGCD\cite{PACCD}                 & \textbf{63.45}                   & 12.84                 & 25.28                   &                   & 48.98                   & 19.05                 & 18.35                   &                   & \textbf{74.55}                   & 9.32                  & 25.04                   &                   & \underline{34.04}                   & 15.74                 & 19.04                   \\
		PromptCCD\cite{PROMPTCCD}               & 60.04                   & \underline{10.48}                 & \underline{31.87}                   &                   & \underline{52.73}                   & \underline{16.18}                 & \underline{30.04}                   &                   & 71.45                   & \underline{7.85}                  & \underline{34.12}                   &                   & 31.73                   & \underline{12.65}                 & \textbf{24.82}                   \\
		Ours                    & \underline{61.55}                   & \textbf{9.65}                  & \textbf{33.26}                   &                   & \textbf{59.76}                   & \textbf{15.07}                 & \textbf{32.73}                   &                   & \underline{72.08}                   & \textbf{7.42}                  & \textbf{41.47}                   &                   & \textbf{39.07}                   & \textbf{10.13}                 & \underline{19.70}                   \\ \bottomrule
	\end{tabular}
  \label{tab:Comparison results all}
\end{table*}

\subsection{Experimental Setups}\label{5.1}
\subsubsection{Datasets} The experiments employ four widely used and challenging fine-grained datasets, including Caltech-UCSD Birds-200-2011 (CUB200)~\cite{CUB}, MIT67~\cite{mit}, Stanford Dogs (Dogs)~\cite{dogs}, and FGVC-Aircraft (Air)~\cite{air}.
Referring to the experimental settings of PA-CGCD~\cite{PACCD} and PromptCCD~\cite{PROMPTCCD}, for each dataset, the first 70\% of classes are known classes, and the remaining 30\% of classes are used as unknown classes. The ratio of training data to test data is 1$:$1.
The training data is divided into $D^l$, $D^u_1$, $D^u_2$, and $D^u_3$ which arrive sequentially. The division of each data set is shown in Table \ref{tab:The number of classes in different training data} and Table \ref{tab:Data splits}.

\subsubsection{Evaluation metric} According to PA-CGCD~\cite{PACCD} and GM~\cite{GMCCD}, three evaluation metrics, namely $\mathcal{M}_\text{o}$, $\mathcal{M}_\text{d}$, and $\mathcal{M}_\text{f}$, are employed.
For each stage $t$, the calculation method of accuracy is defined as the following equation:

\begin{equation}
\mathcal{M}^t = \frac{1}{|D|}\sum^{|D|}_{i=1}\mathbb{I}(y_i=h^*(y^*_i)),
\label{zhunquelv} 
\end{equation}
where $t$ represents the stage. $|D|$ is the size of the test set. $h^*$ is the optimal matching between the pseudo-label and the true label.

$\mathcal{M}_\text{o}$ and $\mathcal{M}_\text{d}$ represent the average accuracy rates of known classes and novel classes in the stages where $t>0$, The calculation formula is as follows:

\begin{equation}
\mathcal{M} = \frac{1}{|T|}\sum_{t\in T}\mathcal{M}^t,
\label{allstagesM} 
\end{equation}
where $|T|$ is the number of stages, and $\mathcal{M}^t$ is the accuracy rate of stage $t$.

$\mathcal{M}_\text{f}$ represents the maximum value of the decline in the accuracy rate of known classes. The calculation formula is the following equation:

\begin{equation}
\mathcal{M}_\text{f} = \max(\mathcal{M}^0_\text{o} - \mathcal{M}^t_\text{o}),
\label{Mf} 
\end{equation}
where $\mathcal{M}^0_\text{o}$ is the recognition accuracy rate of known classes in the initial stage, and $\mathcal{M}^t_\text{o}$ is the recognition accuracy rate of known classes in stage $t$.

\subsubsection{Implementation details} The ResNet-18 pre-trained on ImageNet is used as the backbone.
On all datasets, the number of training epochs in the initial stage is 60, and the number of training epochs in the incremental stage is 30.
For the backbone, the AdamW optimizer is adopted. The initial learning rate is set to 0.0001.
For the projector, the AdamW optimizer is adopted. The learning rates are set to 0.001.
In the $\mathcal{L}_\text{CL}$, $\alpha$ and $\sigma$ are set to 32 and 0.1 respectively.
In the $\mathcal{L}_\text{G}$, $d$ is set to 256.
In fine discovery, $k$ for filtering is set to 10.
The batch size is set to 128.
All experimental results are the averages obtained after multiple runs.
The framework is built with PyTorch on a NVIDIA RTX 3090 GPU\footnote{\url{https://github.com/HaobingLiu/IDOD}}.

\begin{table}[t]
  \centering
  \caption{Estimation of the number of novel classes on CUB200, numbers in parentheses are true numbers of novel classes.}
  \renewcommand{\arraystretch}{1.5}
  \setlength{\tabcolsep}{2pt}
	\begin{tabular}{crrr}
		\toprule
		\textbf{Method}    & \multicolumn{1}{c}{\textbf{Classes (20)}} & \multicolumn{1}{c}{\textbf{Classes (40)}} & \multicolumn{1}{c}{\textbf{Classes (60)}} \\ \hline
		PA-CGCD   & 25.3                               & 62.3                               & 94.7                               \\
		PromptCCD & 23.7                               & 34.3                               & 42.0                               \\
		Ours      & \textbf{18.7}                               & \textbf{37.7}                               & \textbf{65.3}                               \\ \bottomrule
	\end{tabular}
  \label{tab:Comparison results number}
\end{table}

\begin{table}[t]
  \centering
  \caption{Comparison of the required additional space. The measurement of additional space required for IDOD includes the space of static pool and dynamic pool. * indicates that the adopted backbone is ViT. The additional storage space required by IDOD is independent of the backbone, considering that IDOD only stores representations.}
  \renewcommand{\arraystretch}{1.5}
  \renewcommand{\heavyrulewidth}{1pt}
	\begin{tabular}{crrrr}
		\toprule
		\textbf{Method}    & \multicolumn{1}{c}{\textbf{CUB200}} & \multicolumn{1}{c}{\textbf{MIT67}} & \multicolumn{1}{c}{\textbf{Dogs}} & \multicolumn{1}{c}{\textbf{Air}} \\ \hline
		GM        & 78.7M                      & 55.9M                     & 65.2M                    & 61.8M                   \\
		PA-CGCD   & 46.1M                      & 45.8M                     & 45.9M                    & 46.0M                   \\
		PromptCCD* & 360.5M                      & 119.4M                     & 216.3M                    & 180.2M                   \\
		Ours      & \textbf{3.5M}                       & \textbf{5.3M}                      & \textbf{5.4M}                     & \textbf{3.9M}                    \\ \bottomrule
	\end{tabular}
  \label{tab:Addition space}
\end{table}

\subsection{Comparison with State-of-the-arts Methods}\label{5.2}
A series of experiments are conducted to compare the proposed method with state-of-the-arts methods in terms of recognition accuracy, category number estimation and space occupation, as shown in Table~\ref{tab:Comparison results all}, Table~\ref{tab:Comparison results number} and Table~\ref{tab:Addition space}.

\subsubsection{Main Results}
In Table~\ref{tab:Comparison results all}, the proposed method has better recognition performance for novel classes on the CUB200, MIT67 and Dogs datasets than other CCD methods.
Although some recognition performance does not completely surpass other methods,
IDOD not only accurately recognizes novel classes but also maintains the performance for known classes, realizing a better balance between the recognition of known and novel classes.
This is attributed to the fact that the proposed method better handles the feature requirements of novel class discovery and classification, taking into account the diversity and discrimination of features.
The proposed method also surpasses state-of-the-arts methods in preventing forgetting. Considering the effectiveness of orthogonal prototypes and representation replay in preventing forgetting.

\subsubsection{Estimating the Number of Novel Classes}
In Table~\ref{tab:Comparison results number}, the estimation results of the number of novel classes by different methods are presented.
The number of categories estimated by CGCD~\cite{PACCD} is much larger than the true number of categories, while that estimated by PromptCCD~\cite{PROMPTCCD} is much smaller than the true number. The proposed method is closer to the true number of categories.
This is attributed to the fact that the proposed method considers more diverse features, which is conducive to discovering novel classes.
Joint discovery also minimizes the error of discovering novel classes as much as possible, making the final estimated result of the number of novel classes more accurate.
In addition, fine discovery enables IDOD to discover novel classes with a granularity similar to that of known classes, which makes the estimation of the number of novel classes more reasonable.

\subsubsection{Comparison of Additional Storage Space}
In Table~\ref{tab:Addition space}, the additional space required by different methods to deal with catastrophic forgetting is shown. The proposed method significantly reduces the space overhead.
GM~\cite{GMCCD} and CGCD~\cite{PACCD} need to save additional models, and GM also needs to save data samples.
Promptccd~\cite{PROMPTCCD} requires sampling from the GMM pool to prevent forgetting. These methods all require a relatively large amount of storage space.
The proposed method only needs to retain partial sample representations, greatly reducing the space overhead.

\subsection{Ablation Study}\label{5.3}
Ablation experiments are conducted to verify the effectiveness of diversity enhancement, joint discovery, and orthogonal prototypes. The experimental results are shown in Table~\ref{tab:Ablation study}.

\begin{table}[t]
  \centering
  \caption{Ablation experiment results of Independent Enrichment of Diversity Module (IED), Joint Discovery of Novelty Module (JDN), Continuous Increment by Orthogonality Module (CIO) on CUB200.}
  \renewcommand{\arraystretch}{1.5}
  \renewcommand{\heavyrulewidth}{1pt}
	\begin{tabular}{cccrrr}
		\toprule
		\textbf{IED} & \textbf{JDN} & \textbf{CIO} & \multicolumn{1}{c}{\textbf{$\mathcal{M}_\text{o}\uparrow$}} & \multicolumn{1}{c}{\textbf{$\mathcal{M}_\text{f}\downarrow$}} & \multicolumn{1}{c}{\textbf{$\mathcal{M}_\text{d}\uparrow$}} \\ \hline
		\checkmark   &     &     & 44.29                   & 21.93                 & 13.19                   \\
		    & \checkmark   &     & 45.02                   & 21.59                 & 17.03                   \\
		    &     & \checkmark   & 58.16                   & 9.93                  & 10.80                   \\
		    & \checkmark   & \checkmark   & 59.26                   & 12.88                 & 25.04                   \\
		\checkmark   &     & \checkmark   & 61.32                   & 10.39                 & 23.14                   \\
		\checkmark   & \checkmark   &     & 58.35                   & 13.28                 & 27.81                   \\
		\checkmark   & \checkmark   & \checkmark   & \textbf{61.55}                   & \textbf{9.65}                  & \textbf{33.26}                   \\ \bottomrule
	\end{tabular}
  \label{tab:Ablation study}
\end{table}

\subsubsection{Diversity Enrichment}
Diversity enrichment is mainly achieved through the contrastive loss and the independent training of the backbone and the projector.
The design of the ablation experiment is that the prototype is the class mean rather than learnable, and the backbone and the projector are trained together.
The enrichment of diversity improves the performance of discovering novel classes, demonstrating its effectiveness in improving feature diversity.

\subsubsection{Joint Discovery}
Joint discovery converts multi-stage novel class discovery into a single stage, reducing the accumulation of errors.
The design of the ablation experiment is not to save the previous novel class representations. Only retain the two steps of data split and coarse discovery. Joint discovery improves the performance of discovering novel classes.
Considering that joint discovery reduces the error of discovering novel classes, more accurate pseudo-labels are assigned to unlabeled data, enabling the model to learn with more reliable pseudo-labels.

\subsubsection{Orthogonal Prototypes}
Orthogonal prototypes not only obtain discriminative features by making each class orthogonal to each other in the feature space, but also prevents forgetting by reserving the optimal positions for all classes.
The design of the ablation experiment is to train the projector only using cross-entropy. In Table~\ref{tab:Ablation study}, the orthogonal prototypes reduce the degree of forgetting of known classes, verifying its effectiveness in classification and preventing forgetting.

\begin{figure*}[t]
    \centering
    \subfloat[][]{
        \includegraphics[width=0.45\textwidth]{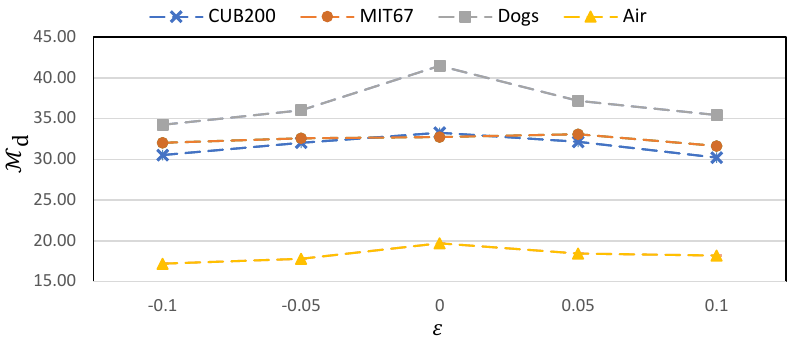}
        \label{fig:plot1}
    }
    \hspace{0.15cm}
    \subfloat[][]{
        \includegraphics[width=0.45\textwidth]{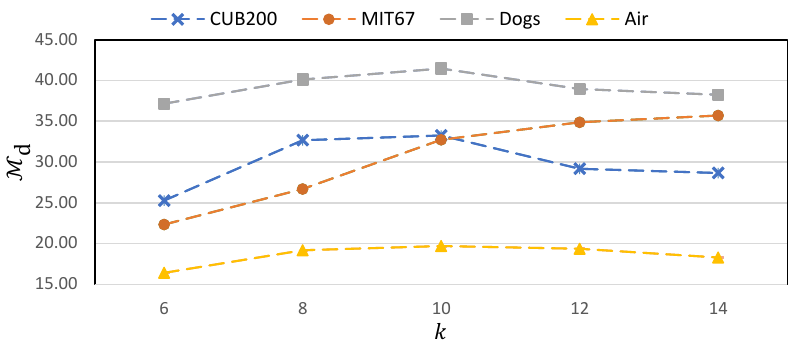}
        \label{fig:plot2}
    }
    \caption{The influence of $\epsilon$ and $k$ on $\mathcal{M}_\text{d}$.}
    \label{para}
\end{figure*}

\subsection{Hyperparameter Analysis}\label{5.4}
The influence of the threshold $\epsilon$ in no-parametric split and that of screening in fine discovery is also analyzed. The experimental results are shown in the Figure~\ref{para}.
On the four datasets, it can be clearly observed that IDOD is robust to the changes of $\epsilon$ and $k$.

Figure~\ref{fig:plot1} shows the impact of $\epsilon$ on $\mathcal{M}_\text{d}$ across four datasets.
The influence of the threshold in no-parametric split affects the results of the preliminary partitioning of novel and known unlabeled data, and ultimately affects the effect of distinguishing known and unknown classes for samples.
The experiment evaluated the performance of the threshold ranging from -0.1 to 0.1. It can be observed that when the threshold is set to 0, the novel class and the known class can be distinguished more accurately. In addition, the final accuracy result is less affected by this threshold, considering that parametric split needs to be performed after non-parametric split, the threshold is finally set to 0.

Figure~\ref{fig:plot2} shows the impact of $k$ on $\mathcal{M}_\text{d}$ across four datasets.
In fine discovery, the value of $k$ used for screening influences the results of novel class discovery. The larger the value of $k$ is, the more samples are retained. However, more unreliable samples may be introduced.
These samples may come from other classes. Considering that the results of coarse discovery are not completely reliable.
A smaller value of $k$ will retain samples that are closer to the cluster center. However, it may lead to too few samples. This can easily restrict the model from learning novel classes. The experiment evaluated the performance of $k$ ranging from 6 to 14. It can be observed that the performance of the method is relatively stable with respect to the value of $k$.

\begin{figure}[t]
    \centering
    \subfloat[CUB200-Backbone]{
        \includegraphics[width=0.2\textwidth]{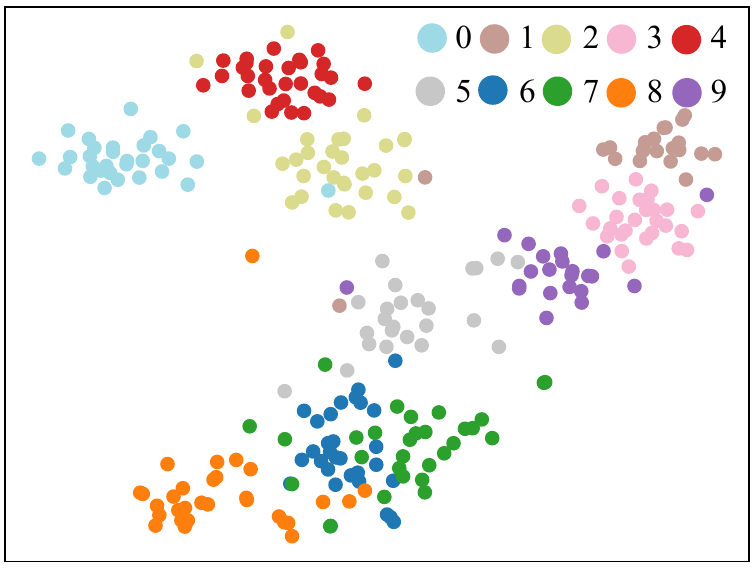}
        \label{fig:plot1.1}
    }
    \hspace{0.5em}
    \subfloat[CUB200-Projector]{
        \includegraphics[width=0.2\textwidth]{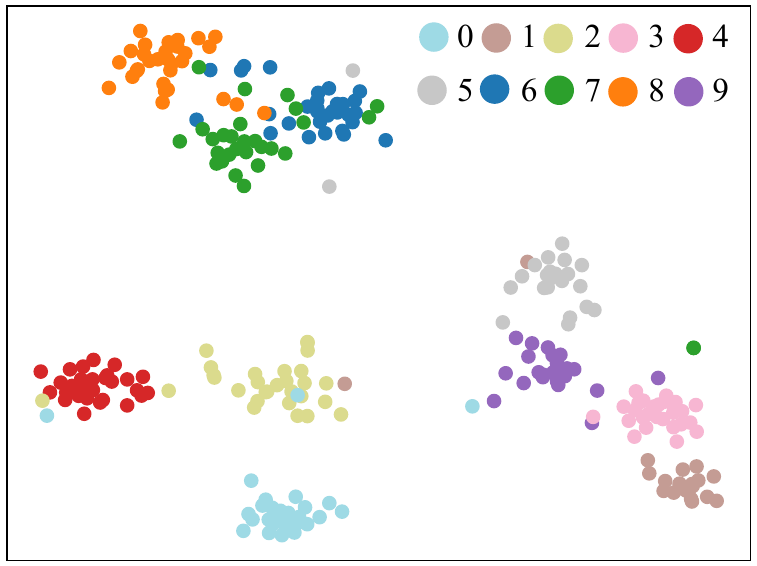}
        \label{fig:plot2.1}
    }
    \hspace{0.5em}
    \subfloat[Dogs-Backbone]{
        \includegraphics[width=0.2\textwidth]{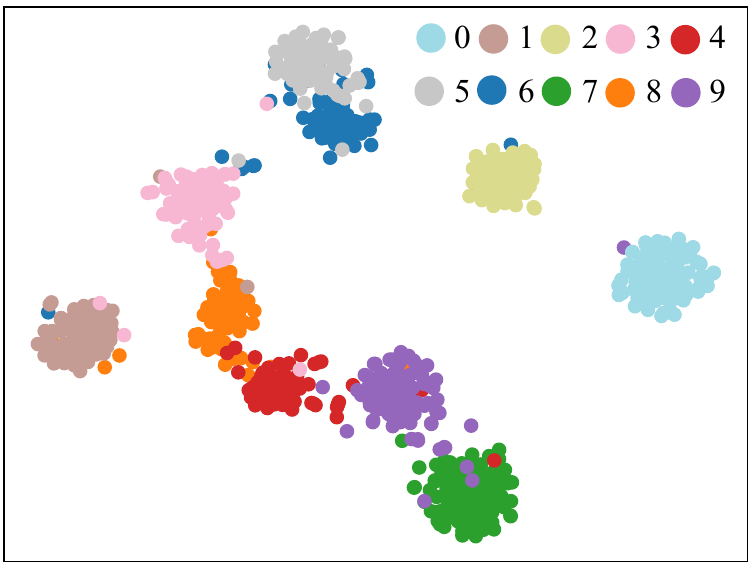}
        \label{fig:plot3.1}
    }
    \hspace{0.5em}
    \subfloat[Dogs-Projector]{
        \includegraphics[width=0.2\textwidth]{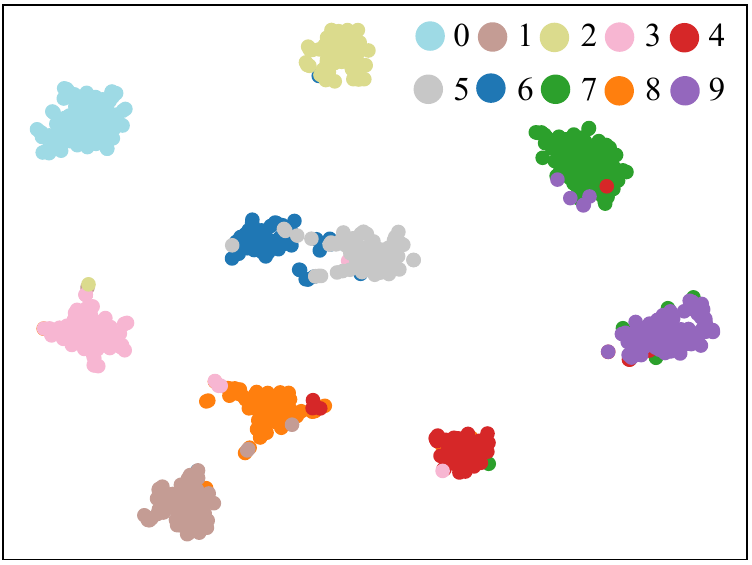}
        \label{fig:plot4.1}
    }
    \caption{Visualization results.}
    \label{visual}
\end{figure}

\subsection{Visualization Results}
To verify the feature extraction capabilities of the backbone and projector, the visualization results on the CUB200 and Dogs datasets are presented.
The visualization results of the representations extracted by the backbone and the projector are shown in Figure~\ref{visual}.

For each dataset, 10 classes are selected for display.
It can be observed that with the assistance of the orthogonal groups, the representation obtained by the projector is more suitable for classification.
The representations of the same class are more compact among themselves, and the spacing between different classes is larger.
The backbone can extract more diverse features to discover novel classes.

\subsection{Detailed Experimental Results}

More detailed experimental results regarding the additional space
overhead and recognition accuracy are presented in Table~\ref{tab:Comparison} and Table~\ref{tab:Comparison results}.
Table~\ref{tab:Comparison} shows the more detailed additional space usage of IDOD.
Table~\ref{tab:Comparison results} more detailed experimental results, which include the estimation of the number of novel classes at each stage, as well as the recognition performance for known and novel classes.

It can be observed that estimation of the number of novel classes is relatively accurate, considering that the joint discovery module reduces the error of discovering novel classes.
As the number of stages increases, the decline in model performance is relatively small, demonstrating the ability to handle multiple arrivals of data.
In addition, the spaces occupied by the static pool and the dynamic pool are similar, and the space overhead is significantly reduced.

\begin{table}[t]
  \centering
  \caption{Detailed additional space usage of IDOD.}
  \renewcommand{\arraystretch}{1.5}
  \renewcommand{\heavyrulewidth}{1pt}
  \label{tab:Comparison}
	\begin{tabular}{crrrr}
		\toprule
		\textbf{Component}    & \multicolumn{1}{c}{\textbf{CUB200}} & \multicolumn{1}{c}{\textbf{MIT67}} & \multicolumn{1}{c}{\textbf{Dogs}} & \multicolumn{1}{c}{\textbf{Air}} \\ \midrule
		Static Pool  & 1.4M                       & 2.7M                      & 3.2M                     & 2.6M                    \\
		Dynamic Pool & 2.1M                       & 2.6M                      & 2.2M                     & 1.3M                    \\ \bottomrule
	\end{tabular}
\end{table}

\begin{table}[t]
  \centering
  \caption{Detailed experimental results on four datasets.}
  \renewcommand{\arraystretch}{1.5}
  \renewcommand{\heavyrulewidth}{1pt}
  \setlength{\tabcolsep}{2pt}
  \label{tab:Comparison results}
	\begin{tabular}{ccrrrcccrrr}
            \toprule
		\multirow{2}{*}{\textbf{Stage}} & \multirow{2}{*}{} & \multicolumn{3}{c}{\textbf{CUB200}}                                            & \multirow{2}{*}{}    & \multirow{2}{*}{\textbf{Stage}} & \multirow{2}{*}{}    & \multicolumn{3}{c}{\textbf{MIT67}}                                             \\ \cline{3-5} \cline{9-11}
		                       &                   & \multicolumn{1}{c}{$n$} & \multicolumn{1}{c}{$\mathcal{M}_\text{o}$} & \multicolumn{1}{c}{$\mathcal{M}_\text{d}$} &                      &                        &                      & \multicolumn{1}{c}{$n$} & \multicolumn{1}{c}{$\mathcal{M}_\text{o}$} & \multicolumn{1}{c}{$\mathcal{M}_\text{d}$} \\ \hline
		$t=1~(20)$                      &                   & 18.7                     & 64.37                     & 36.64                     &                      & $t=1~(7)$                      &                      & 17.3                     & 65.52                     & 34.67                     \\
		$t=2~(40)$                      &                   & 37.7                    & 62.32                     & 35.12                     & \multicolumn{1}{l}{} & $t=2~(14)$                      & \multicolumn{1}{l}{} & 26.3                     & 57.41                     & 31.45                     \\
		$t=3~(60)$                      &                   & 65.3                     & 61.12                     & 28.03                     & \multicolumn{1}{l}{} & $t=3~(21)$                      & \multicolumn{1}{l}{} & 35.3                     & 56.37                     & 32.08                     \\ \hline
		\multirow{2}{*}{\textbf{Stage}} & \multirow{2}{*}{} & \multicolumn{3}{c}{\textbf{Dogs}}                                              & \multirow{2}{*}{}    & \multirow{2}{*}{\textbf{Stage}} & \multirow{2}{*}{}    & \multicolumn{3}{c}{\textbf{Air}}                                               \\ \cline{3-5} \cline{9-11}
		                       &                   & \multicolumn{1}{c}{$n$} & \multicolumn{1}{c}{$\mathcal{M}_\text{o}$} & \multicolumn{1}{c}{$\mathcal{M}_\text{d}$} &                      &                        &                      & \multicolumn{1}{c}{$n$} & \multicolumn{1}{c}{$\mathcal{M}_\text{o}$} & \multicolumn{1}{c}{$\mathcal{M}_\text{d}$} \\ \hline
		$t=1~(12)$                      & \multirow{3}{*}{} & 14.7                     & 74.12                     & 40.46                     & \multirow{3}{*}{}    & $t=1~(10)$                      & \multirow{3}{*}{}    &15.0                     & 42.15                    & 21.02                     \\
		$t=2~(24)$                      &                   & 28.7                     & 71.52                     & 43.89                     &                      & $t=2~(20)$                      &                      & 28.7                     & 38.82                     & 19.30                     \\
		$t=3~(36)$                      &                   & 43.0                     & 70.62                     & 40.08                     &                      & $t=3~(30)$                      &                      & 33.3                     & 36.24                     & 18.78                     \\ \bottomrule
	\end{tabular}
\end{table}

\section{Conclusion}
In this paper, IDOD is proposed, which can achieve a better balance between novel class discovery and classification in CCD.
IDOD makes the backbone and projector focus on diverse features and discriminative features respectively by independent enrichment of diversity module and orthogonal prototypes.
IDOD also contains a joint discovery module that assigns more accurate pseudo-labels to unlabeled data.
Extensive experiments have demonstrated that IDOD outperforms existing methods in achieving a better balance between discovering novel classes and classification.
In future work, it is planned to improve IDOD by adopting better data split method and clustering method.

\begin{acks}
This research is supported in part by the National Science Foundation of China (No. 62302469, No. 62176243), the Natural Science Foundation of Shandong Province (ZR2022QF050, ZR2023QF100), and the Fundamental Research Funds for the Central Universities (202513026).
\end{acks}

\bibliographystyle{ACM-Reference-Format}
\balance
\bibliography{sample-base}


\begin{thebibliography}{46}


\ifx \showCODEN    \undefined \def \showCODEN     #1{\unskip}     \fi
\ifx \showISBNx    \undefined \def \showISBNx     #1{\unskip}     \fi
\ifx \showISBNxiii \undefined \def \showISBNxiii  #1{\unskip}     \fi
\ifx \showISSN     \undefined \def \showISSN      #1{\unskip}     \fi
\ifx \showLCCN     \undefined \def \showLCCN      #1{\unskip}     \fi
\ifx \shownote     \undefined \def \shownote      #1{#1}          \fi
\ifx \showarticletitle \undefined \def \showarticletitle #1{#1}   \fi
\ifx \showURL      \undefined \def \showURL       {\relax}        \fi
\providecommand\bibfield[2]{#2}
\providecommand\bibinfo[2]{#2}
\providecommand\natexlab[1]{#1}
\providecommand\showeprint[2][]{arXiv:#2}

\bibitem[Archana and Jeevaraj(2024)]%
        {deepintro}
\bibfield{author}{\bibinfo{person}{R Archana} {and} \bibinfo{person}{PS~Eliahim Jeevaraj}.} \bibinfo{year}{2024}\natexlab{}.
\newblock \showarticletitle{Deep learning models for digital image processing: a review}.
\newblock \bibinfo{journal}{\emph{Artificial Intelligence Review}} \bibinfo{volume}{57}, \bibinfo{number}{1} (\bibinfo{year}{2024}), \bibinfo{pages}{11}.
\newblock


\bibitem[Zhang et~al\mbox{.}(2024)]%
        {cvcikm}
\bibfield{author}{\bibinfo{person}{Feng Zhang}, \bibinfo{person}{Wei Chen}, \bibinfo{person}{Fei Ding}, \bibinfo{person}{Tengjiao Wang}, \bibinfo{person}{Dawei Lu}, {and} \bibinfo{person}{Jiabin Zheng}.} \bibinfo{year}{2024}\natexlab{}.
\newblock \showarticletitle{Meta-Prompt Tuning Vision-Language Model for Multi-Label Few-Shot Image Recognition}. In \bibinfo{booktitle}{\emph{ACM International Conference on Information and Knowledge Management}}. \bibinfo{pages}{4258--4262}.
\newblock


\bibitem[Zhou(2022)]%
        {introopen}
\bibfield{author}{\bibinfo{person}{Zhi-Hua Zhou}.} \bibinfo{year}{2022}\natexlab{}.
\newblock \showarticletitle{Open-environment machine learning}.
\newblock \bibinfo{journal}{\emph{National Science Review}} \bibinfo{volume}{9}, \bibinfo{number}{8} (\bibinfo{year}{2022}), \bibinfo{pages}{nwac123}.
\newblock


\bibitem[Kejriwal et~al\mbox{.}(2024)]%
        {openworld}
\bibfield{author}{\bibinfo{person}{Mayank Kejriwal}, \bibinfo{person}{Eric Kildebeck}, \bibinfo{person}{Robert Steininger}, {and} \bibinfo{person}{Abhinav Shrivastava}.} \bibinfo{year}{2024}\natexlab{}.
\newblock \showarticletitle{Challenges, evaluation and opportunities for open-world learning}.
\newblock \bibinfo{journal}{\emph{Nature Machine Intelligence}} \bibinfo{volume}{6}, \bibinfo{number}{6} (\bibinfo{year}{2024}), \bibinfo{pages}{580--588}.
\newblock


\bibitem[Parmar et~al\mbox{.}(2023)]%
        {open2}
\bibfield{author}{\bibinfo{person}{Jitendra Parmar}, \bibinfo{person}{Satyendra Chouhan}, \bibinfo{person}{Vaskar Raychoudhury}, {and} \bibinfo{person}{Santosh Rathore}.} \bibinfo{year}{2023}\natexlab{}.
\newblock \showarticletitle{Open-world machine learning: applications, challenges, and opportunities}.
\newblock \bibinfo{journal}{\emph{Comput. Surveys}} \bibinfo{volume}{55}, \bibinfo{number}{10} (\bibinfo{year}{2023}), \bibinfo{pages}{1--37}.
\newblock


\bibitem[Mundt et~al\mbox{.}(2023)]%
        {open3}
\bibfield{author}{\bibinfo{person}{Martin Mundt}, \bibinfo{person}{Yongwon Hong}, \bibinfo{person}{Iuliia Pliushch}, {and} \bibinfo{person}{Visvanathan Ramesh}.} \bibinfo{year}{2023}\natexlab{}.
\newblock \showarticletitle{A wholistic view of continual learning with deep neural networks: Forgotten lessons and the bridge to active and open world learning}.
\newblock \bibinfo{journal}{\emph{Neural Networks}}  \bibinfo{volume}{160} (\bibinfo{year}{2023}), \bibinfo{pages}{306--336}.
\newblock


\bibitem[Wang et~al\mbox{.}(2023)]%
        {opencikm}
\bibfield{author}{\bibinfo{person}{Tianle Wang}, \bibinfo{person}{Zihan Wang}, \bibinfo{person}{Weitang Liu}, {and} \bibinfo{person}{Jingbo Shang}.} \bibinfo{year}{2023}\natexlab{}.
\newblock \showarticletitle{Wot-class: Weakly supervised open-world text classification}. In \bibinfo{booktitle}{\emph{ACM International Conference on Information and Knowledge Management}}. \bibinfo{pages}{2666--2675}.
\newblock


\bibitem[Liu et~al\mbox{.}(2023)]%
        {IJCAI-NCD}
\bibfield{author}{\bibinfo{person}{Jiaming Liu}, \bibinfo{person}{Yangqiming Wang}, \bibinfo{person}{Tongze Zhang}, \bibinfo{person}{Yulu Fan}, \bibinfo{person}{Qinli Yang}, {and} \bibinfo{person}{Junming Shao}.} \bibinfo{year}{2023}\natexlab{}.
\newblock \showarticletitle{Open-world semi-supervised novel class discovery}. In \bibinfo{booktitle}{\emph{International Joint Conference on Artificial Intelligence}}. \bibinfo{pages}{4002--4010}.
\newblock


\bibitem[Wang et~al\mbox{.}(2024)]%
        {introNCD}
\bibfield{author}{\bibinfo{person}{Weishuai Wang}, \bibinfo{person}{Ting Lei}, \bibinfo{person}{Qingchao Chen}, {and} \bibinfo{person}{Yang Liu}.} \bibinfo{year}{2024}\natexlab{}.
\newblock \showarticletitle{Semantic-Guided Novel Category Discovery}. In \bibinfo{booktitle}{\emph{AAAI Conference on Artificial Intelligence}}, Vol.~\bibinfo{volume}{38}. \bibinfo{pages}{5607--5614}.
\newblock


\bibitem[Zhang et~al\mbox{.}(2022)]%
        {GMCCD}
\bibfield{author}{\bibinfo{person}{Xinwei Zhang}, \bibinfo{person}{Jianwen Jiang}, \bibinfo{person}{Yutong Feng}, \bibinfo{person}{Zhi-Fan Wu}, \bibinfo{person}{Xibin Zhao}, \bibinfo{person}{Hai Wan}, \bibinfo{person}{Mingqian Tang}, \bibinfo{person}{Rong Jin}, {and} \bibinfo{person}{Yue Gao}.} \bibinfo{year}{2022}\natexlab{}.
\newblock \showarticletitle{Grow and merge: A unified framework for continuous categories discovery}.
\newblock \bibinfo{journal}{\emph{Advances in Neural Information Processing Systems}}  \bibinfo{volume}{35} (\bibinfo{year}{2022}), \bibinfo{pages}{27455--27468}.
\newblock


\bibitem[Zhao et~al\mbox{.}(2025)]%
        {CGCD2025}
\bibfield{author}{\bibinfo{person}{Zihao Zhao}, \bibinfo{person}{Xiao Li}, \bibinfo{person}{Zhonghao Chang}, {and} \bibinfo{person}{Ningge Hu}.} \bibinfo{year}{2025}\natexlab{}.
\newblock \showarticletitle{Multi-view contrastive learning with maximal mutual information for continual generalized category discovery}.
\newblock \bibinfo{journal}{\emph{Expert Systems with Applications}}  \bibinfo{volume}{266} (\bibinfo{year}{2025}), \bibinfo{pages}{125994}.
\newblock


\bibitem[Oh et~al\mbox{.}(2024)]%
        {closer}
\bibfield{author}{\bibinfo{person}{Junghun Oh}, \bibinfo{person}{Sungyong Baik}, {and} \bibinfo{person}{Kyoung~Mu Lee}.} \bibinfo{year}{2024}\natexlab{}.
\newblock \showarticletitle{CLOSER: Towards Better Representation Learning for Few-Shot Class-Incremental Learning}. In \bibinfo{booktitle}{\emph{European Conference on Computer Vision}}. \bibinfo{pages}{18--35}.
\newblock


\bibitem[Kim et~al\mbox{.}(2023)]%
        {PACCD}
\bibfield{author}{\bibinfo{person}{Hyungmin Kim}, \bibinfo{person}{Sungho Suh}, \bibinfo{person}{Daehwan Kim}, \bibinfo{person}{Daun Jeong}, \bibinfo{person}{Hansang Cho}, {and} \bibinfo{person}{Junmo Kim}.} \bibinfo{year}{2023}\natexlab{}.
\newblock \showarticletitle{Proxy anchor-based unsupervised learning for continuous generalized category discovery}. In \bibinfo{booktitle}{\emph{IEEE/CVF International Conference on Computer Vision}}. \bibinfo{pages}{16688--16697}.
\newblock


\bibitem[Tian et~al\mbox{.}(2024)]%
        {space}
\bibfield{author}{\bibinfo{person}{Songsong Tian}, \bibinfo{person}{Lusi Li}, \bibinfo{person}{Weijun Li}, \bibinfo{person}{Hang Ran}, \bibinfo{person}{Xin Ning}, {and} \bibinfo{person}{Prayag Tiwari}.} \bibinfo{year}{2024}\natexlab{}.
\newblock \showarticletitle{A survey on few-shot class-incremental learning}.
\newblock \bibinfo{journal}{\emph{Neural Networks}}  \bibinfo{volume}{169} (\bibinfo{year}{2024}), \bibinfo{pages}{307--324}.
\newblock


\bibitem[Pian et~al\mbox{.}(2023)]%
        {CILaddition1}
\bibfield{author}{\bibinfo{person}{Weiguo Pian}, \bibinfo{person}{Shentong Mo}, \bibinfo{person}{Yunhui Guo}, {and} \bibinfo{person}{Yapeng Tian}.} \bibinfo{year}{2023}\natexlab{}.
\newblock \showarticletitle{Audio-visual class-incremental learning}. In \bibinfo{booktitle}{\emph{IEEE/CVF International Conference on Computer Vision}}. \bibinfo{pages}{7799--7811}.
\newblock


\bibitem[Zhou et~al\mbox{.}(2024)]%
        {CIL}
\bibfield{author}{\bibinfo{person}{Da-Wei Zhou}, \bibinfo{person}{Qi-Wei Wang}, \bibinfo{person}{Zhi-Hong Qi}, \bibinfo{person}{Han-Jia Ye}, \bibinfo{person}{De-Chuan Zhan}, {and} \bibinfo{person}{Ziwei Liu}.} \bibinfo{year}{2024}\natexlab{}.
\newblock \showarticletitle{Class-incremental learning: A survey}.
\newblock \bibinfo{journal}{\emph{IEEE Transactions on Pattern Analysis and Machine Intelligence}} (\bibinfo{year}{2024}).
\newblock


\bibitem[Wu et~al\mbox{.}(2022)]%
        {introCIL}
\bibfield{author}{\bibinfo{person}{Tz-Ying Wu}, \bibinfo{person}{Gurumurthy Swaminathan}, \bibinfo{person}{Zhizhong Li}, \bibinfo{person}{Avinash Ravichandran}, \bibinfo{person}{Nuno Vasconcelos}, \bibinfo{person}{Rahul Bhotika}, {and} \bibinfo{person}{Stefano Soatto}.} \bibinfo{year}{2022}\natexlab{}.
\newblock \showarticletitle{Class-incremental learning with strong pre-trained models}. In \bibinfo{booktitle}{\emph{IEEE/CVF Conference on Computer Vision and Pattern Recognition}}. \bibinfo{pages}{9601--9610}.
\newblock


\bibitem[Hersche et~al\mbox{.}(2022)]%
        {CILKNOWLEDGE1}
\bibfield{author}{\bibinfo{person}{Michael Hersche}, \bibinfo{person}{Geethan Karunaratne}, \bibinfo{person}{Giovanni Cherubini}, \bibinfo{person}{Luca Benini}, \bibinfo{person}{Abu Sebastian}, {and} \bibinfo{person}{Abbas Rahimi}.} \bibinfo{year}{2022}\natexlab{}.
\newblock \showarticletitle{Constrained few-shot class-incremental learning}. In \bibinfo{booktitle}{\emph{Proceedings of the IEEE/CVF conference on computer vision and pattern recognition}}. \bibinfo{pages}{9057--9067}.
\newblock


\bibitem[Li and Hoiem(2017)]%
        {CILKNOWLEDGE2}
\bibfield{author}{\bibinfo{person}{Zhizhong Li} {and} \bibinfo{person}{Derek Hoiem}.} \bibinfo{year}{2017}\natexlab{}.
\newblock \showarticletitle{Learning without forgetting}.
\newblock \bibinfo{journal}{\emph{IEEE transactions on pattern analysis and machine intelligence}} \bibinfo{volume}{40}, \bibinfo{number}{12} (\bibinfo{year}{2017}), \bibinfo{pages}{2935--2947}.
\newblock


\bibitem[Rebuffi et~al\mbox{.}(2017)]%
        {CILKNOWLEDGE3}
\bibfield{author}{\bibinfo{person}{Sylvestre-Alvise Rebuffi}, \bibinfo{person}{Alexander Kolesnikov}, \bibinfo{person}{Georg Sperl}, {and} \bibinfo{person}{Christoph~H Lampert}.} \bibinfo{year}{2017}\natexlab{}.
\newblock \showarticletitle{icarl: Incremental classifier and representation learning}. In \bibinfo{booktitle}{\emph{Proceedings of the IEEE conference on Computer Vision and Pattern Recognition}}. \bibinfo{pages}{2001--2010}.
\newblock


\bibitem[Wu et~al\mbox{.}(2019)]%
        {CILKNOWLEDGE4}
\bibfield{author}{\bibinfo{person}{Yue Wu}, \bibinfo{person}{Yinpeng Chen}, \bibinfo{person}{Lijuan Wang}, \bibinfo{person}{Yuancheng Ye}, \bibinfo{person}{Zicheng Liu}, \bibinfo{person}{Yandong Guo}, {and} \bibinfo{person}{Yun Fu}.} \bibinfo{year}{2019}\natexlab{}.
\newblock \showarticletitle{Large scale incremental learning}. In \bibinfo{booktitle}{\emph{Proceedings of the IEEE/CVF conference on computer vision and pattern recognition}}. \bibinfo{pages}{374--382}.
\newblock


\bibitem[Belouadah and Popescu(2019)]%
        {CILDATA1}
\bibfield{author}{\bibinfo{person}{Eden Belouadah} {and} \bibinfo{person}{Adrian Popescu}.} \bibinfo{year}{2019}\natexlab{}.
\newblock \showarticletitle{Il2m: Class incremental learning with dual memory}. In \bibinfo{booktitle}{\emph{Proceedings of the IEEE/CVF international conference on computer vision}}. \bibinfo{pages}{583--592}.
\newblock


\bibitem[Castro et~al\mbox{.}(2018)]%
        {CILDATA2}
\bibfield{author}{\bibinfo{person}{Francisco~M Castro}, \bibinfo{person}{Manuel~J Mar{\'\i}n-Jim{\'e}nez}, \bibinfo{person}{Nicol{\'a}s Guil}, \bibinfo{person}{Cordelia Schmid}, {and} \bibinfo{person}{Karteek Alahari}.} \bibinfo{year}{2018}\natexlab{}.
\newblock \showarticletitle{End-to-end incremental learning}. In \bibinfo{booktitle}{\emph{Proceedings of the European conference on computer vision (ECCV)}}. \bibinfo{pages}{233--248}.
\newblock


\bibitem[Hou et~al\mbox{.}(2019)]%
        {CILDATA3}
\bibfield{author}{\bibinfo{person}{Saihui Hou}, \bibinfo{person}{Xinyu Pan}, \bibinfo{person}{Chen~Change Loy}, \bibinfo{person}{Zilei Wang}, {and} \bibinfo{person}{Dahua Lin}.} \bibinfo{year}{2019}\natexlab{}.
\newblock \showarticletitle{Learning a unified classifier incrementally via rebalancing}. In \bibinfo{booktitle}{\emph{Proceedings of the IEEE/CVF conference on computer vision and pattern recognition}}. \bibinfo{pages}{831--839}.
\newblock


\bibitem[Zhu et~al\mbox{.}(2021)]%
        {CILDATA4}
\bibfield{author}{\bibinfo{person}{Fei Zhu}, \bibinfo{person}{Xu-Yao Zhang}, \bibinfo{person}{Chuang Wang}, \bibinfo{person}{Fei Yin}, {and} \bibinfo{person}{Cheng-Lin Liu}.} \bibinfo{year}{2021}\natexlab{}.
\newblock \showarticletitle{Prototype augmentation and self-supervision for incremental learning}. In \bibinfo{booktitle}{\emph{Proceedings of the IEEE/CVF Conference on Computer Vision and Pattern Recognition}}. \bibinfo{pages}{5871--5880}.
\newblock


\bibitem[Chaudhry et~al\mbox{.}(2018)]%
        {CILWEI1}
\bibfield{author}{\bibinfo{person}{Arslan Chaudhry}, \bibinfo{person}{Puneet~K Dokania}, \bibinfo{person}{Thalaiyasingam Ajanthan}, {and} \bibinfo{person}{Philip~HS Torr}.} \bibinfo{year}{2018}\natexlab{}.
\newblock \showarticletitle{Riemannian walk for incremental learning: Understanding forgetting and intransigence}. In \bibinfo{booktitle}{\emph{Proceedings of the European conference on computer vision (ECCV)}}. \bibinfo{pages}{532--547}.
\newblock


\bibitem[Schwarz et~al\mbox{.}(2018)]%
        {CILWEI2}
\bibfield{author}{\bibinfo{person}{Jonathan Schwarz}, \bibinfo{person}{Wojciech Czarnecki}, \bibinfo{person}{Jelena Luketina}, \bibinfo{person}{Agnieszka Grabska-Barwinska}, \bibinfo{person}{Yee~Whye Teh}, \bibinfo{person}{Razvan Pascanu}, {and} \bibinfo{person}{Raia Hadsell}.} \bibinfo{year}{2018}\natexlab{}.
\newblock \showarticletitle{Progress \& compress: A scalable framework for continual learning}. In \bibinfo{booktitle}{\emph{International conference on machine learning}}. PMLR, \bibinfo{pages}{4528--4537}.
\newblock


\bibitem[Ahmed et~al\mbox{.}(2024)]%
        {orco}
\bibfield{author}{\bibinfo{person}{Noor Ahmed}, \bibinfo{person}{Anna Kukleva}, {and} \bibinfo{person}{Bernt Schiele}.} \bibinfo{year}{2024}\natexlab{}.
\newblock \showarticletitle{OrCo: Towards Better Generalization via Orthogonality and Contrast for Few-Shot Class-Incremental Learning}. In \bibinfo{booktitle}{\emph{IEEE/CVF Conference on Computer Vision and Pattern Recognition}}. \bibinfo{pages}{28762--28771}.
\newblock


\bibitem[Zhao et~al\mbox{.}(2023)]%
        {FSCIL}
\bibfield{author}{\bibinfo{person}{Linglan Zhao}, \bibinfo{person}{Jing Lu}, \bibinfo{person}{Yunlu Xu}, \bibinfo{person}{Zhanzhan Cheng}, \bibinfo{person}{Dashan Guo}, \bibinfo{person}{Yi Niu}, {and} \bibinfo{person}{Xiangzhong Fang}.} \bibinfo{year}{2023}\natexlab{}.
\newblock \showarticletitle{Few-shot class-incremental learning via class-aware bilateral distillation}. In \bibinfo{booktitle}{\emph{IEEE/CVF conference on computer vision and pattern recognition}}. \bibinfo{pages}{11838--11847}.
\newblock


\bibitem[Zhang et~al\mbox{.}(2023)]%
        {NCDaddition1}
\bibfield{author}{\bibinfo{person}{Sheng Zhang}, \bibinfo{person}{Salman Khan}, \bibinfo{person}{Zhiqiang Shen}, \bibinfo{person}{Muzammal Naseer}, \bibinfo{person}{Guangyi Chen}, {and} \bibinfo{person}{Fahad~Shahbaz Khan}.} \bibinfo{year}{2023}\natexlab{}.
\newblock \showarticletitle{Promptcal: Contrastive affinity learning via auxiliary prompts for generalized novel category discovery}. In \bibinfo{booktitle}{\emph{IEEE/CVF Conference on Computer Vision and Pattern Recognition}}. \bibinfo{pages}{3479--3488}.
\newblock


\bibitem[An et~al\mbox{.}(2023)]%
        {NCDaddition2}
\bibfield{author}{\bibinfo{person}{Wenbin An}, \bibinfo{person}{Feng Tian}, \bibinfo{person}{Qinghua Zheng}, \bibinfo{person}{Wei Ding}, \bibinfo{person}{QianYing Wang}, {and} \bibinfo{person}{Ping Chen}.} \bibinfo{year}{2023}\natexlab{}.
\newblock \showarticletitle{Generalized category discovery with decoupled prototypical network}. In \bibinfo{booktitle}{\emph{AAAI Conference on Artificial Intelligence}}, Vol.~\bibinfo{volume}{37}. \bibinfo{pages}{12527--12535}.
\newblock


\bibitem[Roy et~al\mbox{.}(2022)]%
        {CILNCD}
\bibfield{author}{\bibinfo{person}{Subhankar Roy}, \bibinfo{person}{Mingxuan Liu}, \bibinfo{person}{Zhun Zhong}, \bibinfo{person}{Nicu Sebe}, {and} \bibinfo{person}{Elisa Ricci}.} \bibinfo{year}{2022}\natexlab{}.
\newblock \showarticletitle{Class-incremental novel class discovery}. In \bibinfo{booktitle}{\emph{European Conference on Computer Vision}}. \bibinfo{pages}{317--333}.
\newblock


\bibitem[Han et~al\mbox{.}(2021)]%
        {NCD11}
\bibfield{author}{\bibinfo{person}{Kai Han}, \bibinfo{person}{Sylvestre-Alvise Rebuffi}, \bibinfo{person}{Sebastien Ehrhardt}, \bibinfo{person}{Andrea Vedaldi}, {and} \bibinfo{person}{Andrew Zisserman}.} \bibinfo{year}{2021}\natexlab{}.
\newblock \showarticletitle{Autonovel: Automatically discovering and learning novel visual categories}.
\newblock \bibinfo{journal}{\emph{IEEE Transactions on Pattern Analysis and Machine Intelligence}} \bibinfo{volume}{44}, \bibinfo{number}{10} (\bibinfo{year}{2021}), \bibinfo{pages}{6767--6781}.
\newblock


\bibitem[Jia et~al\mbox{.}(2021)]%
        {NCD12}
\bibfield{author}{\bibinfo{person}{Xuhui Jia}, \bibinfo{person}{Kai Han}, \bibinfo{person}{Yukun Zhu}, {and} \bibinfo{person}{Bradley Green}.} \bibinfo{year}{2021}\natexlab{}.
\newblock \showarticletitle{Joint representation learning and novel category discovery on single-and multi-modal data}. In \bibinfo{booktitle}{\emph{Proceedings of the IEEE/CVF International Conference on Computer Vision}}. \bibinfo{pages}{610--619}.
\newblock


\bibitem[Zhao and Han(2021)]%
        {NCD13}
\bibfield{author}{\bibinfo{person}{Bingchen Zhao} {and} \bibinfo{person}{Kai Han}.} \bibinfo{year}{2021}\natexlab{}.
\newblock \showarticletitle{Novel visual category discovery with dual ranking statistics and mutual knowledge distillation}.
\newblock \bibinfo{journal}{\emph{Advances in Neural Information Processing Systems}}  \bibinfo{volume}{34} (\bibinfo{year}{2021}), \bibinfo{pages}{22982--22994}.
\newblock


\bibitem[Zhong et~al\mbox{.}(2021)]%
        {NCD14}
\bibfield{author}{\bibinfo{person}{Zhun Zhong}, \bibinfo{person}{Linchao Zhu}, \bibinfo{person}{Zhiming Luo}, \bibinfo{person}{Shaozi Li}, \bibinfo{person}{Yi Yang}, {and} \bibinfo{person}{Nicu Sebe}.} \bibinfo{year}{2021}\natexlab{}.
\newblock \showarticletitle{Openmix: Reviving known knowledge for discovering novel visual categories in an open world}. In \bibinfo{booktitle}{\emph{Proceedings of the IEEE/CVF Conference on Computer Vision and Pattern Recognition}}. \bibinfo{pages}{9462--9470}.
\newblock


\bibitem[Vaze et~al\mbox{.}(2022)]%
        {NCD21}
\bibfield{author}{\bibinfo{person}{Sagar Vaze}, \bibinfo{person}{Kai Han}, \bibinfo{person}{Andrea Vedaldi}, {and} \bibinfo{person}{Andrew Zisserman}.} \bibinfo{year}{2022}\natexlab{}.
\newblock \showarticletitle{Generalized category discovery}. In \bibinfo{booktitle}{\emph{Proceedings of the IEEE/CVF Conference on Computer Vision and Pattern Recognition}}. \bibinfo{pages}{7492--7501}.
\newblock


\bibitem[Wen et~al\mbox{.}(2023)]%
        {NCD22}
\bibfield{author}{\bibinfo{person}{Xin Wen}, \bibinfo{person}{Bingchen Zhao}, {and} \bibinfo{person}{Xiaojuan Qi}.} \bibinfo{year}{2023}\natexlab{}.
\newblock \showarticletitle{Parametric classification for generalized category discovery: A baseline study}. In \bibinfo{booktitle}{\emph{Proceedings of the IEEE/CVF International Conference on Computer Vision}}. \bibinfo{pages}{16590--16600}.
\newblock


\bibitem[Vaze et~al\mbox{.}(2024)]%
        {NCD23}
\bibfield{author}{\bibinfo{person}{Sagar Vaze}, \bibinfo{person}{Andrea Vedaldi}, {and} \bibinfo{person}{Andrew Zisserman}.} \bibinfo{year}{2024}\natexlab{}.
\newblock \showarticletitle{No representation rules them all in category discovery}.
\newblock \bibinfo{journal}{\emph{Advances in Neural Information Processing Systems}}  \bibinfo{volume}{36} (\bibinfo{year}{2024}).
\newblock


\bibitem[Wu et~al\mbox{.}(2023)]%
        {METACCD}
\bibfield{author}{\bibinfo{person}{Yanan Wu}, \bibinfo{person}{Zhixiang Chi}, \bibinfo{person}{Yang Wang}, {and} \bibinfo{person}{Songhe Feng}.} \bibinfo{year}{2023}\natexlab{}.
\newblock \showarticletitle{Metagcd: Learning to continually learn in generalized category discovery}. In \bibinfo{booktitle}{\emph{Proceedings of the IEEE/CVF International Conference on Computer Vision}}. \bibinfo{pages}{1655--1665}.
\newblock


\bibitem[Cendra et~al\mbox{.}(2024)]%
        {PROMPTCCD}
\bibfield{author}{\bibinfo{person}{Fernando~Julio Cendra}, \bibinfo{person}{Bingchen Zhao}, {and} \bibinfo{person}{Kai Han}.} \bibinfo{year}{2024}\natexlab{}.
\newblock \showarticletitle{Promptccd: Learning gaussian mixture prompt pool for continual category discovery}. In \bibinfo{booktitle}{\emph{European Conference on Computer Vision}}. Springer, \bibinfo{pages}{188--205}.
\newblock


\bibitem[Frey and Dueck(2007)]%
        {AP}
\bibfield{author}{\bibinfo{person}{Brendan~J Frey} {and} \bibinfo{person}{Delbert Dueck}.} \bibinfo{year}{2007}\natexlab{}.
\newblock \showarticletitle{Clustering by passing messages between data points}.
\newblock \bibinfo{journal}{\emph{science}} \bibinfo{volume}{315}, \bibinfo{number}{5814} (\bibinfo{year}{2007}), \bibinfo{pages}{972--976}.
\newblock


\bibitem[Wah et~al\mbox{.}(2011)]%
        {CUB}
\bibfield{author}{\bibinfo{person}{Catherine Wah}, \bibinfo{person}{Steve Branson}, \bibinfo{person}{Peter Welinder}, \bibinfo{person}{Pietro Perona}, {and} \bibinfo{person}{Serge Belongie}.} \bibinfo{year}{2011}\natexlab{}.
\newblock \showarticletitle{The caltech-ucsd birds-200-2011 dataset}.
\newblock  (\bibinfo{year}{2011}).
\newblock


\bibitem[Quattoni and Torralba(2009)]%
        {mit}
\bibfield{author}{\bibinfo{person}{Ariadna Quattoni} {and} \bibinfo{person}{Antonio Torralba}.} \bibinfo{year}{2009}\natexlab{}.
\newblock \showarticletitle{Recognizing indoor scenes}. In \bibinfo{booktitle}{\emph{IEEE conference on computer vision and pattern recognition}}. \bibinfo{pages}{413--420}.
\newblock


\bibitem[Khosla et~al\mbox{.}(2011)]%
        {dogs}
\bibfield{author}{\bibinfo{person}{Aditya Khosla}, \bibinfo{person}{Nityananda Jayadevaprakash}, \bibinfo{person}{Bangpeng Yao}, {and} \bibinfo{person}{Fei-Fei Li}.} \bibinfo{year}{2011}\natexlab{}.
\newblock \showarticletitle{Novel dataset for fine-grained image categorization: Stanford dogs}. In \bibinfo{booktitle}{\emph{CVPR workshop on fine-grained visual categorization (FGVC)}}.
\newblock


\bibitem[Maji et~al\mbox{.}(2013)]%
        {air}
\bibfield{author}{\bibinfo{person}{Subhransu Maji}, \bibinfo{person}{Esa Rahtu}, \bibinfo{person}{Juho Kannala}, \bibinfo{person}{Matthew Blaschko}, {and} \bibinfo{person}{Andrea Vedaldi}.} \bibinfo{year}{2013}\natexlab{}.
\newblock \showarticletitle{Fine-grained visual classification of aircraft}.
\newblock \bibinfo{journal}{\emph{arXiv preprint arXiv:1306.5151}} (\bibinfo{year}{2013}).
\newblock


\end{thebibliography}

\end{document}